\documentclass[10pt,twocolumn,letterpaper]{article}

\usepackage[pagenumbers]{cvpr} % To force page numbers, e.g. for an arXiv version

\usepackage[dvipsnames]{xcolor}

\definecolor{cvprblue}{rgb}{0.21,0.49,0.74}
\usepackage[pagebackref,breaklinks,colorlinks,citecolor=cvprblue]{hyperref}

\def\paperID{56} % *** Enter the Paper ID here
\def\CodalabID{: earth-insights} % *** Enter the Codalab Account name here
\def\confName{CVPR}
\def\confYear{2024}

\title{Class Similarity Transition: Decoupling Class Similarities and Imbalance from Generalized Few-shot Segmentation}

\makeatletter
\def\thanks#1{\protected@xdef\@thanks{\@thanks
        \protect\footnotetext{#1}}}
\makeatother

\author{Shihong Wang$^{1*}$, Ruixun Liu$^{1*}$, Kaiyu Li$^{1*}$, Jiawei Jiang$^{2}$, Xiangyong Cao$^{1\dag}$ \thanks{$^*$Equal contribution. \dag Corresponding author.} \\
$^1$ Xi’an Jiaotong University, China. ~$^2$ Sun Yat-Sen University, China.\\
{\tt\small jack3shihong@gmail.com, liuruixun6343@gmail.com, likyoo.ai@gmail.com} \\
{\tt\small jiangjw26@mail2.sysu.edu.cn, caoxiangyong@mail.xjtu.edu.cn} \\
\url{https://github.com/earth-insights/ClassTrans}
}

\begin{document}
\maketitle
\begin{abstract}
% This paper proposes a novel Generalized Few-shot Segmentation (GFSS) framework that addresses the similarity of base classes and novel classes and the imbalance between the support set and the training set. Specifically, we first propose a probability Transition Matrix to capture the similarity. Then, we introduce a Label-Distribution-Aware Margin (LDAM) loss and transductive inference to address the problem of class imbalance as well as overfitting in an end-to-end training paradigm. In addition, by extending the probability Transition Matrix, the proposed method can mitigate the Catastrophic Forgetting of base classes when learning novel classes. With a simple training phase, our proposed method can be applied on top of any segmentation network trained on base classes. We tested our methods on the adapted version of OpenEarthMap. Compared to existing GFSS baselines, our method excels them all from 3\% to 7\%. With stronger backbones, our method can consistently enhance their performance. Additionally, we conducted various ablation studies to validate our claims and demonstrate the effectiveness of our methods. At the completion of this paper, our method ranks second in the OpenEarthMap Land Cover Mapping Few-Shot Challenge without bells and whistles.
In Generalized Few-shot Segmentation (GFSS), a model is trained with a large set of base class samples and then adapted on limited samples of novel classes. This paper focuses on the relevance between base and novel classes, and improves GFSS in two aspects: 1) mining the similarity between base and novel classes to promote the learning of novel classes, and 2) mitigating the class imbalance issue caused by the volume difference between the support set and the training set. Specifically, we first propose a similarity transition matrix to guide the learning of novel classes with base class knowledge. Then, we leverage the Label-Distribution-Aware Margin (LDAM) loss and Transductive Inference to the GFSS task to address the problem of class imbalance as well as overfitting the support set. In addition, by extending the probability transition matrix, the proposed method can mitigate the catastrophic forgetting of base classes when learning novel classes. With a simple training phase, our proposed method can be applied to any segmentation network trained on base classes. We validated our methods on the adapted version of OpenEarthMap. Compared to existing GFSS baselines, our method excels them all from 3\% to 7\% and ranks second in the OpenEarthMap Land Cover Mapping Few-Shot Challenge at the completion of this paper.
% without bells and whistles.
\end{abstract}    
\section{Introduction}
\label{sec:intro}

\begin{figure}[t]
        \centering
        \includegraphics[width=8cm]{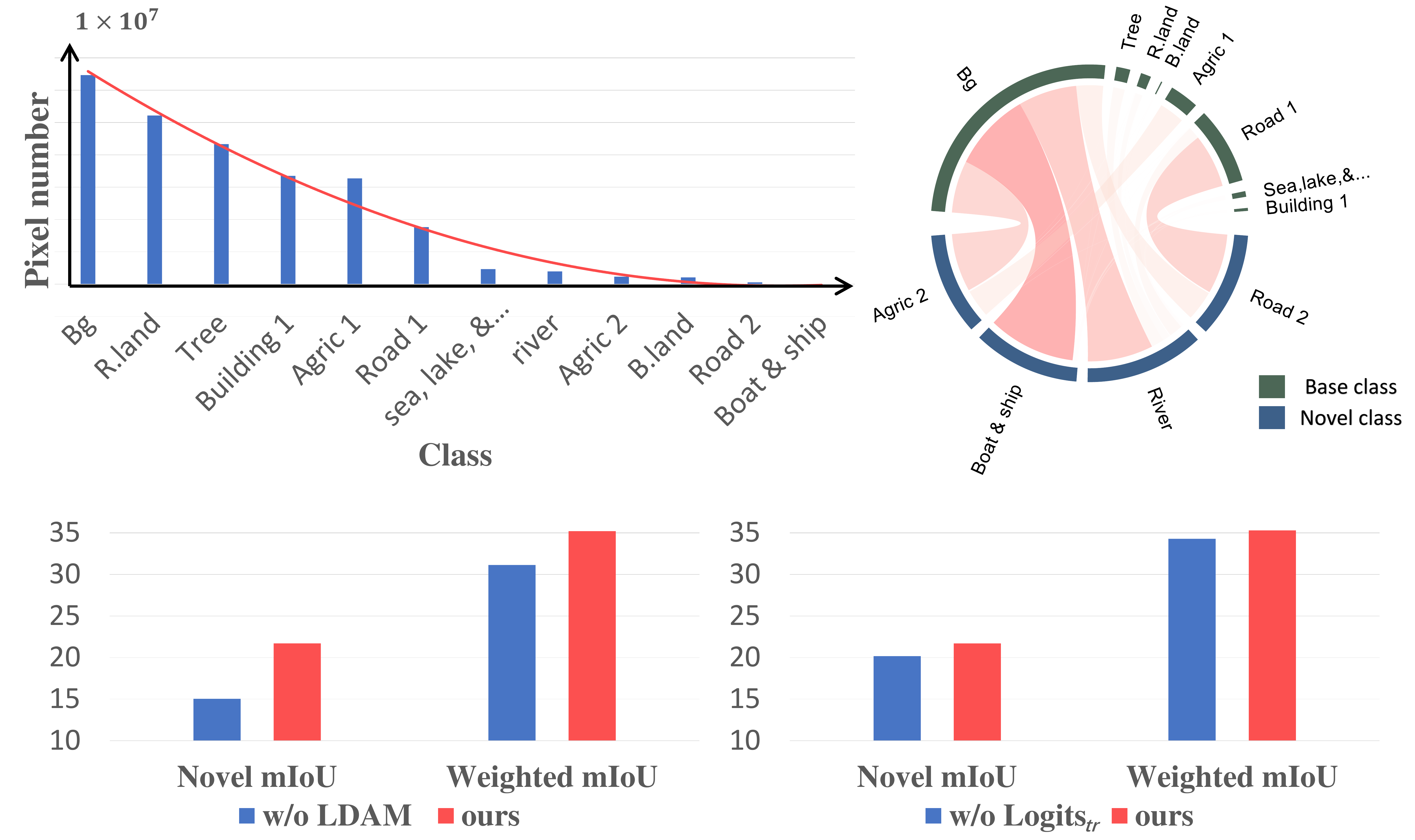}
        \caption{\textbf{The motivation of our proposed method and the experiment result.} \textbf{A)} The top left figure reveals that datasets from the real world exhibit long-tailed distribution. \ie severe class imbalance. The dataset shown here is from \cite{datasetsoem} whose adaptation is used in the challenge. \textbf{B)} The top right figure shows the class relevance by counting the novel class pixels from the support set misclassified by the model after the training phase. The wider and deeper the bond is, the more pixels are wrongly classified as base classes.  \textbf{C)} The bottom figure shows the effectiveness of our proposed components in tackling class imbalance and class similarity problems.}
        \label{fig:intro}
\end{figure}

In recent years, deep learning has revolutionized the field of computer vision, enabling significant advancements in various vision tasks such as image classification \cite{alextnet, resnet}, object detection \cite{rcnn}, and semantic segmentation \cite{unet}. Semantic segmentation, in particular, plays a crucial role in understanding the visual world by assigning a class label to each pixel in an image \cite{sssurvey}. This fine-grained pixel-level prediction facilitates scene understanding and serves as a fundamental component in various applications, including autonomous driving, medical image analysis, and augmented reality. Despite the impressive achievements of deep learning models in semantic segmentation, these approaches still have some limitations. One significant drawback is the heavy reliance on large annotated datasets for training. The process of manually labeling pixel-level annotations for training data is labor-intensive, time-consuming, and costly \cite{coco}, undermining the application of deep learning. 

% Moreover, traditional semantic segmentation models trained on a predefined set of classes struggle to generalize to new classes once deployed. These drawbacks hinder the flexibility and robustness of semantic segmentation systems in real-world scenarios.

Recent efforts have introduced the concept of Few-shot Semantic Segmentation (FSS) \cite{fss-1, fss-2, fss-3}. The goal of FSS is to enable the model to quickly learn to segment novel classes from limited samples, thus mitigating the demand for data annotation. In the FSS paradigm, the model is first trained on a well-annotated dataset comprising only base examples ($\sim$1000x). Then, during the testing phase, a support set consisting of a few examples ($\sim$10x) with novel classes is provided for few-shot learning algorithms. Finally, the performance of FSS algorithms is evaluated on a query set. However, the FSS task only focuses on the performance of novel classes while ignoring base classes, thus leading to the performance decrease in learned base classes,
% 这里要加一个实验，FSS在旧类上的性能在小样本学习之后会变差（可以在第一页右上角添加一个柱状图说明）
% \textcolor{red}{As shown in Fig.~\cite{}}, 
% As , current FSS methods have a performance decrease in learned base classes, 
\ie FSS methods gain fast learning ability by sacrificing the performance of learned knowledge \cite{diam}. But in reality, we would like the neural network to have the same learning ability as humans: quickly learning novel classes from limited samples while preserving what we have learned before.

In recent years, generalized few-shot semantic segmentation (GFSS) \cite{capl, diam} has been proposed to address the shortcomings of FSS setups. GFSS evaluates the model in both base and novel classes after a few shots of learning. Moreover, GFSS relaxes the constraints that FSS has on the data, allowing novel classes to appear in the training set as background and only providing annotations of novel classes in the support set during few-shot learning. In this way, GFSS can emulate human learning behavior in that humans only focus on classifying interested classes when learning for the first time while treating others as background, and humans then need to learn some novel classes from background and treat the remaining classes as the background again. This can effectively alleviate the cost of annotating data for only novel classes that are needed for the second learning.

Although these GFSS works \cite{capl, bam, diam} have shown promising progress, two key elements have been overlooked. The first neglected element is \textit{\textbf{class similarity}}. Previous work has achieved the classification of novel classes by learning prototypes \cite{bam} or a linear classifier \cite{diam} for novel classes during the few shot learning phase, but has overlooked the possibility that novel classes may be relevant to base classes. As shown in Fig. \ref{fig:intro}, relevant classes can share similar features, such as road type 1 in the base class and road type 2 in the novel class. The features of oceans learned in the base-class learning phase can help enhance the classification performance of the novel classes, thereby achieving better results. The second ignored element is \textit{\textbf{class imbalance}}. As shown in Fig. \ref{fig:intro}, in real-world applications, training examples always exhibit a long-tailed distribution, that is, a class imbalance issue exists in the datasets. Such imbalanced data sets challenge learning algorithms by making the model easily biased toward head classes \cite{longtailsurvey}. Essentially, given a large-scale training set and disproportionately small support sets, the GFSS model faces an even more imbalanced data set, while previous work \cite{diam, capl} ignored this essential issue existing in the GFSS task.

% To address the aforementioned issues, this paper adopts the divide-and-conquer strategy and modifies the GFSS method from three aspects in the few-shot learning phase, \ie, maintaining knowledge of base classes, handling class imbalance, and utilizing base-class-to-novel-class similarity. Specifically, \textcolor{red}{describe the three aspects in detail!!!}. 

To address the aforementioned issues, this paper adopts the divide-and-conquer strategy and modifies the GFSS method from three points of view in the few-shot learning phase. Specifically, for utilizing base-class-to-novel-class similarity, we build a similarity transition matrix to guide the learning from base classes to novel classes. The transition matrix is then extended by a base-to-base projection, and an implicit diagonal matrix is learned to encourage maintaining knowledge of the base class. To address the class imbalance between base and novel classes, the  Label-Distribution-Aware Margin (LDAM) loss \cite{longtailldam} with class distribution prior is introduced, which amplifies the effect of sparse novel classes.

The contributions of our work are summarized as:

\begin{itemize}
    \item We determine two key problems that exist in the GFSS task: class similarity and class imbalance between the novel classes and base classes, and accordingly propose the probability transfer matrix and introduce a logit adjustment loss.

    \item We show that our method exceeds the state-of-the-art GFSS methods and validate the effectiveness of the proposed components through a series of ablation and visualization analyses. Our method ranks second in the development phase of the OpenEarthMap challenge without bells and whistles.
    
    % We determine two key problems that exist in the GFSS task that the previous GFSS methods ignored: addressing the class imbalance and utilizing the class similarity between the novel classes and base classes. With careful handling of these two aspects, our determined problems help enlighten the future works of the GFSS task, leading to further advancements in addressing GFSS tasks.
    
    % \item Based on our findings, we propose a novel method that excels current baselines to tackle GFSS, that also can be applied to current segmentation models, armoring them with GFSS capability. Our method gains prominent improvements in performance. Besides, we conducted a series of ablation studies verifying the effectiveness and the strong generalization capability to the backbone design of our method, \ie a stronger backbone can improve the performance of our method further.

    \item Our experiments further illustrate the importance of preventing overfitting on the support set and retaining knowledge of base classes. The issues we emphasized point to the necessary means for advancing GFSS tasks in future research.

\end{itemize}
%-------------------------------------------------------------------------

\section{Related Works}

\noindent \paragraph{Generalized Few-shot Segmentation.} 
Some works recently extended existing FSS methods to GFSS settings: \textbf{A) Prototypical Network.} Instead of learning features as traditional convolution networks do, prototypical networks try to learn a network that outputs the prototypes of the classes. Then given a query image, the prototypical network classifies the pixel by cosine similarity or clustering based on the prototypes of classes and the pixel \cite{prototype1}. Being a pioneering work, Tian \etal \cite{capl} first proposed a prototypical network, named CAPL, for GFSS with two modules to dynamically adapt both base and novel class prototypes to tackle GFSS. Furthermore, \cite{protype3orthogonal} tries to improve the quality of the prototype learned by the feature extractor by orthogonalizing the prototypes while \cite{protype2similarity} addresses the similarities between base and novel classes by a Graph Neural Network with class-contrastive loss. Though some of these works stressed class similarity, none of these methods address the training example imbalances in the training set and support set.  \textbf{B) Unified Classifier with Transductive Inference.} Due to the limited number of novel class examples in the support set, directly learning a classifier for the novel class on the support set will inevitably lead to overfitting, resulting in poor performance on novel classes. \cite{transductive1} introduced Transductive Inference into FSS, \ie a Kullback-Leibler (KL) Divergence term in loss function between the pixel distribution of the model's predictions on query image and that of estimated ground truth pixel distribution, punishing overfitting on support images. Thus training a unified end-to-end classification network is viable. Later, DIaM \cite{diam} extends the idea of transductive learning to GFSS and achieves competitive results. However, DIaM overlooked the utilization of the class similarity. Compared with existing GFSS methods, our proposed method addresses both class similarity and class imbalance, achieving higher performance.

% \noindent \textbf{Learning without Forgetting}

% During the Few-shot Learning phase, directly fine-tuning the classifier obtained from base class learning will eventually lead to significant degradation of the base classes, a phenomenon known as Catastrophic Forgetting. How to deal with Catastrophic Forgetting is the core issue in Continual Learning. Currently, several mainstream methods are proposed to tackle Catastrophic forgetting: \textbf{A) Data Replay} saves some instances of base classes, called exemplar sets. Such exemplar sets are used to overcome catastrophic forgetting when learning novel classes \cite{continualsurvey}. Various methods have been proposed to either 

\noindent \paragraph{Class Similarities.}
Utilizing the knowledge from source classes to aid the classification of target classes, \ie utilizing the knowledge of base classes to help the model learn the novel classes, is known as transfer learning \cite{transfersurvey}. \textbf{A) Parameter controlling} directly controls the parameter of the novel classes, \ie by sharing parameters \cite{transfer1} of the base classes or forcing the parameters of the novel classes to be similar to base classes \cite{transfer2}, to take advantage of learned knowledge. However, parameter controlling does not leverage the similarity effectively. \textbf{B) Adversarial learning} \cite{transfer3dan, transfer4dan} transfers the knowledge from the base model to the new model by learning the model that cannot be discriminated by the discriminator. However, the extra demands to train a discriminator are unrealistic under the GFSS scenario. The exploration of using the similarity between classes to transfer knowledge is a rarely researched field. Our method is a novel method to transfer the knowledge learned from base classes to novel classes.

\noindent \paragraph{Class Imbalance.}
Class-imbalanced learning, also referred as long-tailed learning, aims to learn a classifier that performs well on both head, body, and tail classes (in GFSS setting, base classes, and novel classes). To tackle class imbalance, various methods tackle the problem from different perspectives: \textbf{A) Expert networks} try to learn a shared feature extractor and class-specific classifiers, called expert, to classify head, body, and tail classes \cite{longtailexpert1}. Further, determining which expert to use to obtain the prediction is also studied \cite{longtailrouting}. However, with more parameters to train, the optimization of the expert networks is very data-hungry, especially when limited novel class examples are available in the case of GFSS; \textbf{B) Re-weighting.} Finding that the norm of classification weights for tail classes is disproportionately smaller than that of head classes, decreasing the possibilities for tail classes after Softmax, some researchers \cite{longtailreweight1, longtailreweight2} decided to enhance the performance of classifiers on tail classes by normalizing the weights of linear classifiers by class. However, a recent study \cite{longtaillogits} points out that the correlation between weight norms and classes is dependent on the design of the classifier and the choice of optimizer thus re-weighting strategy fits to specific classifiers and lacks robustness; \textbf{C) Logit adjustment} aims to rectify the logits of the classifier, resulting in a greater punishment for misclassifying tail classes \cite{longtailldam, longtaillogits} by the prior distribution of classes. Further, some methods \cite{longtailunknown1, longtailunknown2} are proposed to estimate the prior when it is unavailable. In our methods, we adopt logit adjustment to mitigate class imbalance in GFSS for it brings nearly no extra overhead of optimizing and is robust to classifier design with no extra demand for training examples..

\begin{figure*}[t]
        \centering
        \includegraphics[height=8cm]{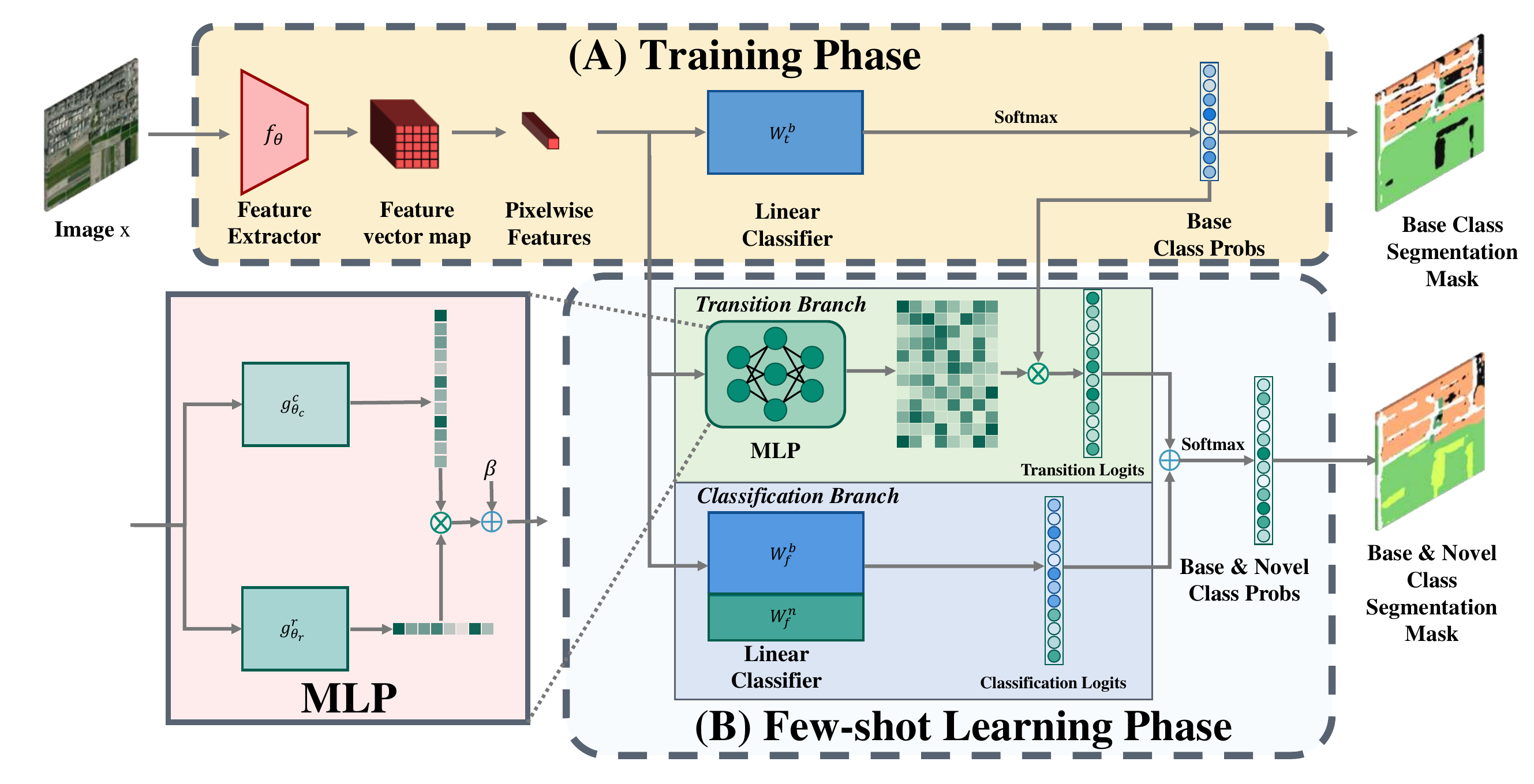}
        \caption{\textbf{Pipeline and our proposed framework.} $\otimes$ refers to matrix product operation, $\oplus$ refers to element-wise addition. The pipeline consists of two phases: the training phase and the few-shot learning Phase. \textbf{A)} First, the training phase learns a shared feature extractor $f_\theta$ and classification weights $W^b_t$ for the base classes, as shown in the figure. Besides, novel classes are treated as background during the training phase. \textbf{B)} Then, the few-shot learning phase learns a MLP that predicts the transition matrix and classification weights $W^b_f, W^n_f$ for the base classes and novel classes respectively. \textbf{C)} With $W^b_t, W^b_f, W^n_f$, and MLP learned, our proposed method gives the transition and classification logits from \textit{Transition Branch} and \textit{Classification Branch} respectively. By merging the transition logits and classification logits, the model gives the final predictions of both base and novel classes on a given image.}
        \label{figure:framework}
\end{figure*}

\section{Methods}

\subsection{Notations}

\noindent\textbf{Preliminaries.} Denote $H$ and $W$ as the height and width of the image $\mathbf{x}\in \mathbb{R}^{|\Omega|\times 3}$, and $\Omega= \{(m,n)\}, m=0,...,H-1, n=0,...,W-1$ is the pixel coordinates. In all generality, we denote a segmentation model $f$ that combines a feature extractor $\phi$ that outputs a feature vector map $\mathbf{V}=\phi(\mathbf{x})\in\mathbb{R}^{|\Omega|\times F}$ where $F$ denotes the channel of the feature vector and a classifier with weight $W\in\mathbb{R}^{K\times F}$ which takes image $\mathbf{x}$ and output segmentation maps $f(\mathbf{x})=\mathbf{P}\in\{0,1\}^{|\Omega|\times K}$ where $K$ is the number of the classes to predict. For simplicity, we omit the bias of the classifier.

\noindent\textbf{Generalized few-shot segmentation.} In GFSS, two sets of classes are considered: the \textit{base} classes, $\mathcal{C}^b$, and the \textit{novel} classes, $\mathcal{C}^n$, and $\mathcal{C}^b\cap\mathcal{C}^n=\emptyset$. In the training phase, the model learns datasets $D_{train}$ which contain base class and treats novel class as background. At test time, the model is evaluated through a series of tasks. In each task, the model first goes through a few-shot learning phase on given support set $\mathbb{S}=\{\mathbf{x}_i, y_i\}^{|\mathbb{S}|}_{i=1}$, containing a few images, along with their corresponding binary segmentation masks $y_i \in \{0, 1\}^{|\Omega|}$, for some novel class sampled from $\mathcal{C}^n$. After the model quickly learns from the support set $\mathbb{S}$, the model is required to give predictions on an unlabeled image, termed \textit{query} image, of both base and novel classes to test the GFSS learning algorithm, \ie for pixel $j$ we have

\[
\mathbf{P}(j) = \begin{bmatrix}
    {\overbrace{
        \begin{matrix}
            p_0
        \end{matrix}}^{\text{bg}}
    },
    &
    {\overbrace{
        \begin{matrix}
            p_1,\dotsb,p_{|\mathcal{C}^n|}
        \end{matrix}}^{\text{base classes}}
    },
    &
    {\overbrace{
        \begin{matrix}
            p_{|\mathcal{C}^n|+1},\dotsb,p_{|\mathcal{C}^n|+|\mathcal{C}^b|}
        \end{matrix}}^{\text{novel classes}}
    }
\end{bmatrix}^T
\]

\noindent in which bg stands for \textit{background}, $(j)$ represents pixel indexing and we omit pixel index $j$ from the right-hand side for simplicity.

\subsection{Similarity transition}

Our idea of class similarity originates from the idea that by giving an image containing both base and novel classes, the model only trained on base classes would wrongly classify a pixel of the novel class as the similar base class. After learning $D_{train}$, given an image containing both novel and base classes, pixel $j$ of novel class $y_n$ will be wrongly classified as base class $y_b$ by the model, \ie

\begin{equation}
p(y_b|x(j)) = \text{Softmax(}W^b_t\circ f_\theta(\mathbf{x}(j))\text{)} \quad y_b=0,\ldots,|\mathcal{C}^b|
\end{equation}

\noindent where $\circ$ denotes matrix product and $f_\theta$, $W^b_t\in\mathbb{R}^{(1 + |\mathcal{C}^b|)\times F}$, represents parameters of feature extractor and linear classifier after training on $D_{train}$. If we obtain $p(y_n|y_b,\mathbf{x}(j))$, thus we would have 

\begin{equation}
    p(y_n|\mathbf{x}(j))=\sum_{y_b=0}^{\mathcal{C}^b}p(y_n|y_b,\mathbf{x}(j))\cdot p(y_b|\mathbf{x}(j))
\end{equation}

\noindent The possibility $p(y_n|y_b,\mathbf{x}(j))$ means, given a pixel $j$ being wrongly classified as base class $y_b$ by $W^b_t$, the possibility of pixel $j$ belongs to novel class $y_n$. Thus, $p(y_n|y_b,\mathbf{x}(j))$ describes the similarity from base class $y_b$ to novel class $y_n$. For simplicity, we denote
\begin{equation}
    s_{hq} := p(y_n=h|y_b=q, \mathbf{x}(j))
\end{equation}
\noindent and then for a pixel $j$ we have a corresponding $|\mathcal{C}^n|\times(1+|\mathcal{C}^b|)$ matrix ($1$ for background)
\begin{equation*}
    \mathbf{S}(\mathbf{x})=
    \begin{bmatrix}
        s_{|\mathcal{C}^b|+1,0} & s_{|\mathcal{C}^b|+1,1} & \cdots & s_{|\mathcal{C}^b|+1,|\mathcal{C}^b|}\\
        s_{|\mathcal{C}^b|+2,0} & s_{|\mathcal{C}^b|+2,1} & \cdots & s_{|\mathcal{C}^b|+2,|\mathcal{C}^b|}\\
        \vdots & \vdots & \ddots & \vdots\\
        s_{|\mathcal{C}^b|+|\mathcal{C}^n|,0} & s_{|\mathcal{C}^b|+|\mathcal{C}^n|,1} & \cdots & s_{|\mathcal{C}^b|+|\mathcal{C}^n|,|\mathcal{C}^b|}\\
    \end{bmatrix}
\end{equation*}

\noindent Note, for simplicity, we omit pixel indexing $(j)$. We call this matrix Similarity Transition Matrix from base classes to novel classes (or abbreviated as Transition Matrix), since it defines the probability of transiting wrongly classified base classes to novel classes. Besides, the elements of $\mathbf{S}(\mathbf{x})$ are positive and each column sums to 1, for it represents the probability.

Note that conditional probability $p(y_b|\mathbf{x}(j))$ is the wrong classification probability given by the classifier after training on $D_{train}$. Thus, with $p(y_b|\mathbf{x}(j))$ and $\mathbf{S}(\mathbf{x})$ obtained, our network consists of two branches, as shown in Fig. \ref{figure:framework}:

\begin{itemize}
    \item \textit{Classification Branch.} The classification branch is a classifier that takes in the feature vector and outputs classification logits. The implementation of classifiers can be in any form, e.g. cosine similarity-based classifiers, Euclidean distance-based clustering, etc. In our implementation, we simply choose a linear classifier as a vanilla classification network does. Thus we have classification logits:
    \begin{equation}
        \text{Logits}_{cl}=\text{cat}([W^b_f, W^n_f])\circ \phi(\mathbf{x})
    \end{equation}
    in which \textit{cat} represents concatenates operation, $W^b_f, W^n_f$ represents the parameter of the linear classifier of base classes and novel classes to learn during few-shot learning.
    
    \item \textit{Transition Branch.} The transition branch takes in the feature vector and outputs the similarity transition logits. Given similarity transition matrix $\mathbf{S}(\mathbf{x})$ and classifier weights $W^b_t$ after training on $D_{train}$. We can obtain the similarity transition logits (abbreviated as transition logits):
    \begin{equation}
         \text{Logits}_{tr}= \mathbf{S}(\mathbf{x}) \circ \text{Softmax}(W^b_t\circ \phi(\mathbf{x}))
    \end{equation}
    To obtain the similarity transition matrix $\mathbf{S}(\mathbf{x})$ we adopt a Multilayer Perceptron (MLP) $\mathbf{S}(\mathbf{x})=g(\mathbf{x})$ to estimate the transition matrix. Also, to leverage the feature extraction functionality provided by the backbone, in our implementation, we take the feature vector map $\mathbf{V}$ as input rather than raw image $\mathbf{x}$ as input. Finally, the output vector of MLP is resized to the shape of the similarity transition matrix.
\end{itemize}

\paragraph{Reducing the difficulty of optimization.} Further on, increasing the number of parameters improves the model's capacity but raises the risk of overfitting as well. In the proposed MLP, it contains $(\mathcal{C}^n \times (1+\mathcal{C}^b)) \times F$ parameters to optimize which burdens the optimization algorithms a lot. To reduce the difficulty of optimizing MLP, we decompose the MLP into two smaller MLPs, $g^r(\mathbf{x})$ and $g^c(\mathbf{x})$, each outputting the rows and columns of the transition matrix respectively, \ie

$$
\mathbf{S}(\mathbf{x}) = g(\mathbf{x}) = g^c_{\theta_c}(\mathbf{x}) \otimes g^r_{\theta_r}(\mathbf{x}) + \beta
$$

\noindent where $\beta$ is a learnable bias that avoids $\mathbf{S}(\mathbf{x})$ ranks 1, $\theta_c, \theta_r$ are the parameters of $g^c,g^r$ respectively, and $\otimes$ represents outer product operation of two vector. By decomposing, we reduce the parameter of the MLP to $(\mathcal{C}^n + 1 + \mathcal{C}^b) \times F$, in our implementation, form $(4\times8)\times 512$ to $(4 + 8)\times 512$.

To our knowledge, we are the first to utilize the similarity transition matrix in GFSS. There is a similar work that also uses transition matrix \cite{labelhierarchy1}. However,  several key differences exist that support our novelty: \textbf{a)} \cite{labelhierarchy1} is originally proposed to tackle multi-category classification while we focus on GFSS. The essences of the fields studied are different. \textbf{b)} The transition probabilities proposed in \cite{labelhierarchy1} are to evaluate the possibility that a fine class belongs to a coarse class, while we mainly utilize the transition probabilities as a measurement of the similarity between base classes and novel classes. 

\subsection{Exploring class imbalance in-depth}

Natural datasets exhibit long-tailed distribution. Simply adopting cross-entropy as Eq. (\ref{eqn:ce}) in which $z_k$ is the $k$-th logits of the output of the model to ensure high confidence on predictions will eventually lead the model biased towards the dominant classes, \ie base classes.

\begin{equation}
    \label{eqn:ce}
    \mathcal{L}_{ce}(\mathbf{x}, y) = -\log \frac{e^{z_y}}{\sum_{k=0}^{|\mathcal{C}^b| + |\mathcal{C}^n|}e^{z_k}}
\end{equation}

This is because, utilizing traditional cross-entropy, the great number of base class samples causes the models to incur a higher punishment when misclassifying base classes as novel classes, leading to larger gradients \cite{longtailsurvey} to mitigate that. On the contrary, misclassifying novel classes as base classes will not result in a great punishment. Consequently, the optimized model is more inclined to predict all samples as base classes, as shown in our experiments.
% 加一个实验

A proven effective way to mitigate the disproportional punishment of base classes and novel classes is rectifying the logits before cross-entropy \cite{longtaillogits, longtailldam}, \ie, amplifying the loss on novel classes. Given the class distribution $n_k$ for $k$-th class, we use such class distribution to rectify the logits. Following LDAM \cite{longtailldam}, we adopt rectified cross-entropy as:

\begin{equation}
    \label{epn:ldam}
    \mathcal{L}_{\textit{LDAM}}(\mathbf{x}, y) = -\log\frac{e^{z_y-\delta_y}}{e^{z_y-\delta_y} + \sum_{k\neq y}e^{z_k}}
\end{equation}

\noindent where $\delta_k=\frac{C}{n_k^{1/4}}$ and $C$ a hyper-parameter to tune. By doing so, we rectify the logits according to the frequency of the class. As a result, the loss of novel classes is amplified. The rest problem is how to obtain the class distribution. In our implementation, we estimate the class distribution via $D_{train}$ and $D_{support}$.

\subsection{Preserving base knowledge}

\begin{figure}[t]
        \centering
        \includegraphics[width=7cm]{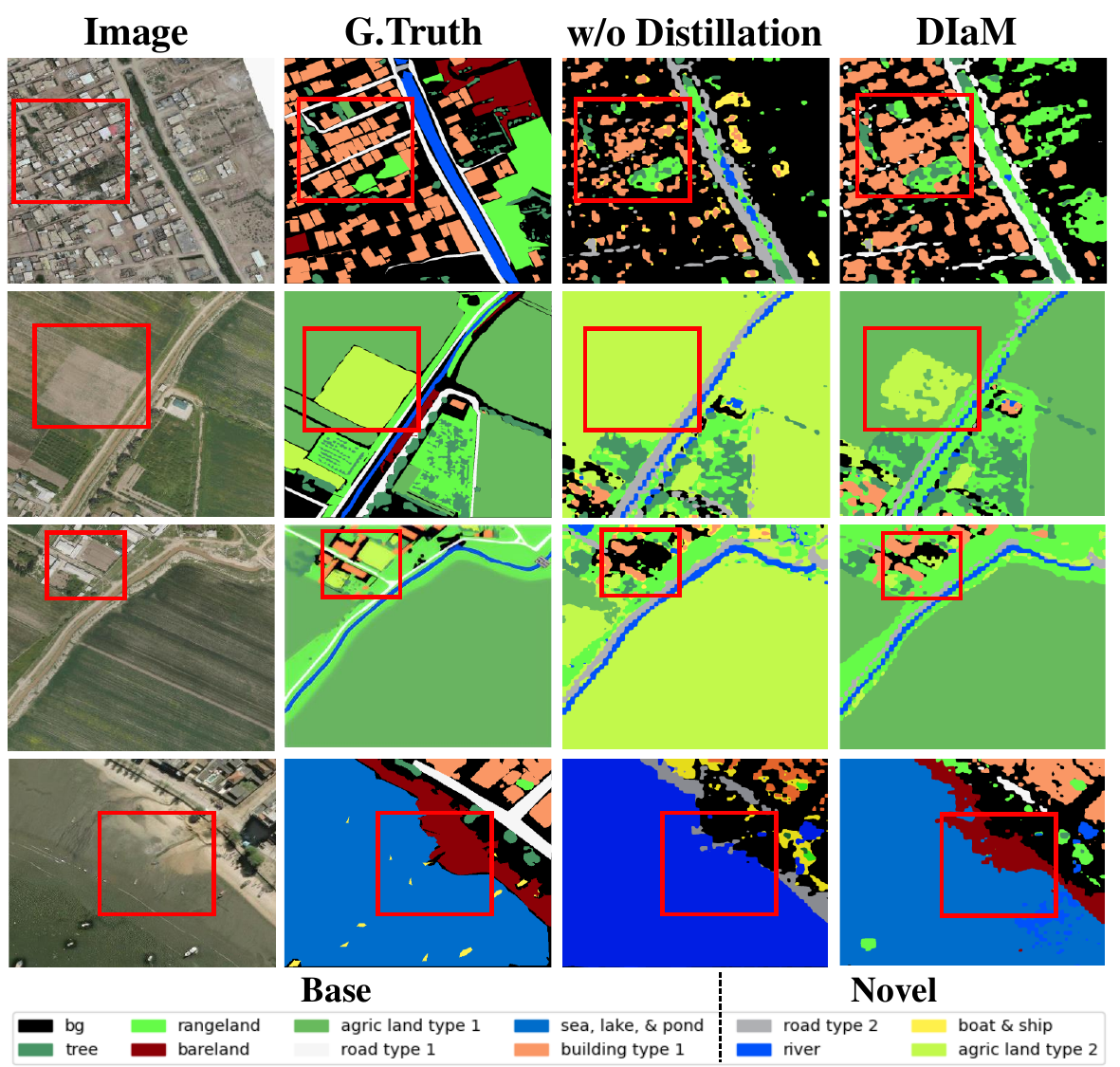}
        \caption{\textbf{Verifying the existence of Catastrophic Forgetting.} G.Truth refers to the ground truth segmentation mask of a given image. ``w/o Distillation'' represents training DIaM \cite{diam} by removing Knowledge Distillation. ``DIaM'' represents our adaptation of DIaM \cite{diam} on our dataset.}
        \label{fig:forgetting}
\end{figure}

The setting of GFSS prevents the supporting images from providing explicit supervision, thus during few-shot learning when continually updating the parameters, the lack of supervision of the base class will eventually result in the model forgetting the knowledge of base classes obtained from the training phase. As shown in Fig. \ref{fig:forgetting}, after removing the Knowledge Distillation, DIaM \cite{diam} wrongly classifies the base class to the novel class, \eg, Agric land type 1 to Agric land type 2, Sea to River, and Building type 1 to the background.

The previous method DIaM \cite{diam} adopts Knowledge Distillation to overcome the forgetting

\begin{equation}
    \mathcal{L}_{KD} = \mathbf{KL}(\pi_{\text{new2old}}(\mathbf{P})||\hat{\mathbf{P}})
\end{equation}

\noindent where $\hat{\mathbf{P}} = W^b_t\cdot\phi(\mathbf{x})$ is the prediction of the base classifier obtained after training on $D_{train}$, $\mathbf{P}$ is the prediction of the classifier for both base and novel during few-shot learning, and $\pi_{\text{new2old}}(\cdot)$ is a projection function which maps novel class logits for during training phase, novel classes are treated as background, \ie

\begin{equation*}
    \pi_{\text{new2old}}(\mathbf{P})(j)=\left[ p_0+\sum_{k=1}^{|\mathcal{C}^n|}p_{|\mathcal{C}^b|+k}, p_1, p_2, \cdots, p_{|\mathcal{C}^b|} \right]^T 
\end{equation*}

In our proposed method, we preserve the knowledge of base classes in a different way by extending the Transition Matrix from $|\mathcal{C}^n|\times(1+|\mathcal{C}^b|)$ to $(1 + |\mathcal{C}^b| + |\mathcal{C}^n|)\times(1+|\mathcal{C}^b|)$, as shown in Eq. (\ref{eqn:extend}), where $\mathbf{S}^{\text{base2novel}}(\mathbf{x})$ is the aforementioned $\mathbf{S}(\mathbf{x})$. Each element in $\mathbf{S}^{\text{base2base}}(\mathbf{x})$ represents the transition probability from base classes to base classes. Under an extreme situation in which the diagonal elements of $\mathbf{S}^{\text{base2base}}(\mathbf{x})$ are close to 1 while others are close to 0, the learned knowledge is preserved after outer product. The detailed exemplification is shown in Sec. \ref{sec:vis_trans}.

\begin{equation}
    \label{eqn:extend}
    \hat{\mathbf{S}}(\mathbf{x})=
    \begin{bmatrix}
        \mathbf{S}^{\text{base2base}}(\mathbf{x})\\
        \mathbf{S}^{\text{base2novel}}(\mathbf{x})
    \end{bmatrix}
\end{equation}

\begin{table*}[htp]
\centering
\caption{\textbf{Results of our methods compared to baselines.} We implement all baselines using ResNet-101 as the backbone to make a fair comparison with our method (Ours-RN). Besides, we report our highest submission results from the competition portal which uses ConvNext-L as the backbone (Ours-CN).}
\label{tab:main}
\resizebox{0.9\textwidth}{!}{%
\begin{tabular}{l|l|cccccccc}
\hline
Method&Backbone&Base&Road type 2&River&Boat \& ship&Agric land type 2&Novel& Average mIoU& Weighted mIoU\\ 
\hline
CAPL& ResNet-101&27.64& \textbf{0.75}& 17.15& 0.07& 1.25& 4.81& 16.02& 13.99\\
BAM& ResNet-101&36.33&     0.00&5.31&\underline{0.45}&8.30&3.52& 19.93& 16.64\\
DIaM& ResNet-101&\underline{37.41}&0.12&5.96&\textbf{1.83}&8.61&4.13& 20.77 & 17.44\\
Ours-RN& ResNet-101& 37.37& 0.04& \underline{18.93}& 0.05& \underline{21.74}& \underline{10.19}& \underline{23.78}&\underline{21.06}\\
\hline
Ours-CN& ConvNext-L & \textbf{55.46}& \underline{0.44}& \textbf{39.13}& 0.00& \textbf{47.28}&\textbf{21.71}& \textbf{38.58}& \textbf{35.21}\\
\hline
\end{tabular}%
}
\end{table*}

\subsection{Preventing from overfitting the support set}

Learning a classifier from $D_{support}$ for epochs leads to the classifier overfitting the support set, as shown in Fig. \ref{fig:overfit1}. To prevent the overfitting, \cite{transductive1} introduces a transductive normalization term. We adapt the proposed method into our framework as

\begin{equation}
    \mathcal{L}_{\pi} = \hat{\mathcal{P}}_Q\log(\frac{\hat{\mathcal{P}}_Q}{\pi})
\end{equation}

\noindent with $\hat{\mathcal{P}}_Q=\frac{1}{|\Omega|}\sum_{k=0}^{1+|\mathcal{C}^b| + |\mathcal{C}^n|}\mathcal{P}_Q$ being the proportion of each class in the model's prediction of the query image and $\pi$ being that of the ground truth. The term $\mathcal{L}_{\pi}$ punishes the model when the proportion of each class in the prediction of query image mismatches that of the ground truth, so as to prevent overfitting on the support set. 

$\pi$ is agnostic during few-shot learning, so we estimate the $\pi$ by the model's prediction, \ie

\begin{equation}
    \pi^t = \left\{
    \begin{array}{ll}
        \hat{\mathcal{P}}_Q^0, &  0 \leq t \leq t_\pi \\
        \hat{\mathcal{P}}_Q^{t_\pi}, & t > t_\pi
    \end{array}
\right.
\end{equation}

\noindent In summary, the final objective is 

\begin{equation}
    \min_{W^b_f,W^n_f,\theta_r,\theta_c}\mathcal{L} = \mathcal{L}_{\textit{LDAM}} + \lambda\cdot\mathcal{L}_\pi
\end{equation}

\noindent where $\lambda$ is a hyper-parameter to tune.

% \begin{figure}[t]
%   \centering
%   \fbox{\rule{0pt}{2in} \rule{0.9\linewidth}{0pt}}
%    %\includegraphics[width=0.8\linewidth]{egfigure.eps}

%    \caption{Category mapping}
%    \label{fig:onecol}
% \end{figure}

\section{Experiments}

\subsection{Experimental setting}

\paragraph{Datasets.} To test the effectiveness of our method we test our methods on an adapted version of OpenEarthMap \cite{datasetsoem}. The adapted version consists of 408 samples of the original OpenEarthMap benchmark dataset and has 15 classes. We select 8 classes as base classes including background, and 4 as novel classes. The support set consists of 20 image-label pair examples by 5 examples for each of the 4 novel classes. The labels for each image in the support set do not contain any of the base classes. Also, in each 5-set example, the labels contain only one novel class. Details of the labels are shown in Tab. \ref{tab:ablation}.

\paragraph{Evaluation protocol.}

We report the mean intersection-over-union (mIoU) over the classes. In our tables, \textit{Base} and \textit{Novel} refer to mIoU over base classes and novel classes, respectively. \textit{Average} refers to mIoU over all classes. Following \cite{diam}, directly compare \textit{Average} mIoU is unfair for novel classes, so we adopt \textit{Weighted} mIoU which weights \textit{Base} mIoU and \textit{Novel} mIoU by 0.6:0.4.

\paragraph{Baselines.} We adapt CAPL \cite{capl} and DIaM \cite{diam} to our dataset and compare the results. Though BAM \cite{bam} is proposed from FSS, we adapt BAM to GFSS following \cite{diam} and report the results in the GFSS task. For all implementations, we train the classifier by 100 epochs. Necessary down-sampling and up-sampling is implemented.

\subsection{Main Results}

\begin{figure}[h]    % 常规操作\begin{figure}开头说明插入图片
  \centering            % 前面说过，图片放置在中间
  \subfloat  % 第一张子图的下标（注意：注释要写在[]中括号内）
  {
      \label{fig:overfit_tl}\includegraphics[width=0.3\textwidth]{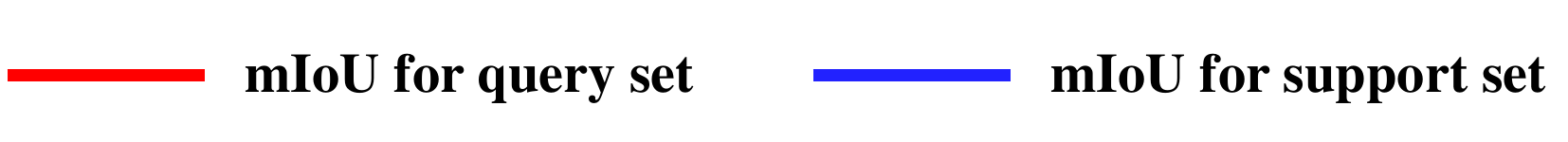}
  }\\
  \setcounter{subfigure}{0}
  \subfloat[$\lambda$=0]   % 第一张子图的下标（注意：注释要写在[]中括号内）
  {
      \label{fig:overfit1}\includegraphics[width=0.14\textwidth]{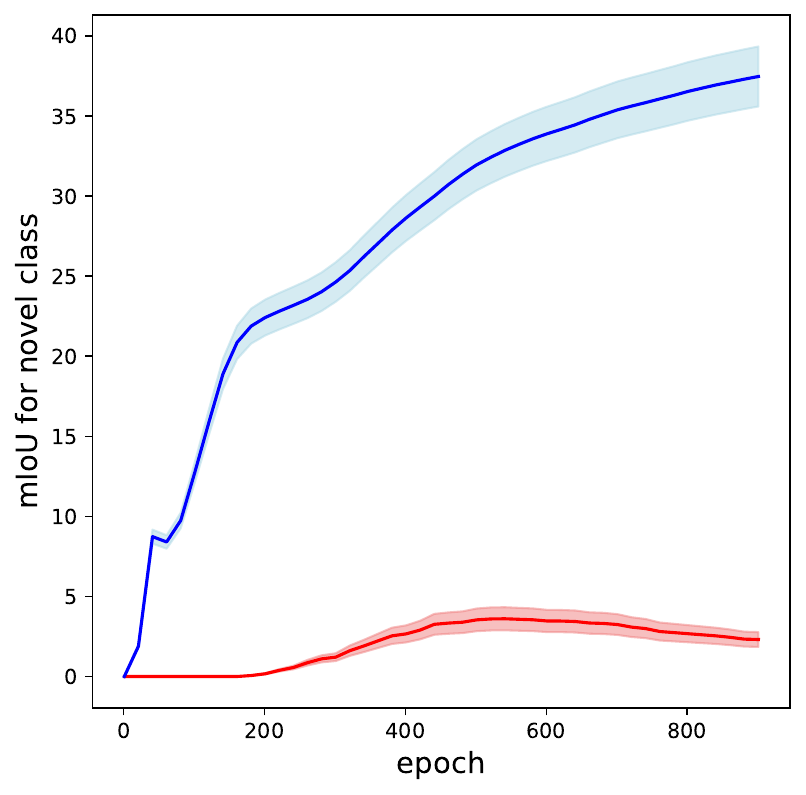}
  }
  \subfloat[$\lambda$=1]
  {
      \label{fig:overfit2}\includegraphics[width=0.14\textwidth]{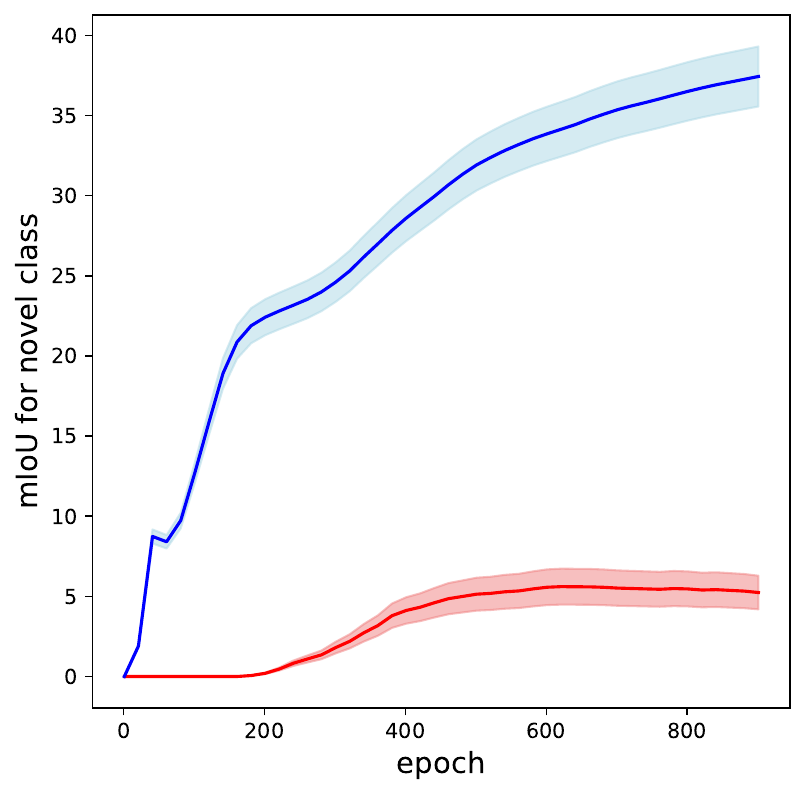}
  }
\subfloat[$\lambda$=4]
  {
      \label{fig:overfit3}\includegraphics[width=0.14\textwidth]{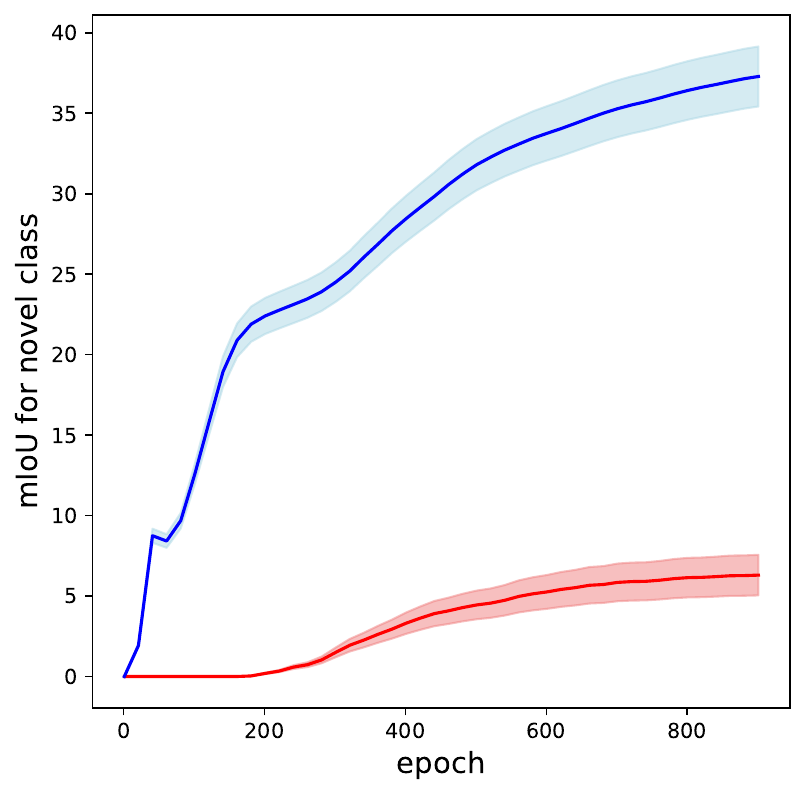}
  }
  \caption{\textbf{The quantitative result of $\mathcal{L}_\pi$.} $\lambda$ represents the weight of $\mathcal{L}_\pi$ in the final objective function. \textbf{a)} $\lambda=0$ indicates the $\mathcal{L}_\pi$ is removed. \textbf{b,c)} $\lambda=1$ and $\lambda=4$ indicates the weight of $\mathcal{L}_\pi$ is set to $1$ and $4$ respectively.}    % 整个图片的说明，注释写在{}内
  \label{fig:overfit}            % 整个图片的标签编号，注意这里跟子图是一样的道理，标签不能重复 
\end{figure}

\textbf{Comparison to state-of-the-art GFSS.} Tab. \ref{tab:main} reports the comparison of our method and current SOTA methods in GFSS. To make a fair comparison, we use ResNet-101 as the backbone for all methods compared. It can be seen that: \textbf{A)} Compared with other methods, our method with ResNet-101 backbone, i.e. Ours-RN, outperforms existing SOTA. And the main improvement comes from the novel classes. \textbf{B)} The great majority of the novel class improvement comes from \textit{River} and \textit{Agric land type 2}, this is because similar base class \textit{Sea} helps the model classify \textit{River}. Similar to \textit{Agric land type 2}.

% 加一个ablation，mask 掉transition matrix中海洋和road type 1，然后查看性能

Further on, to test our method's effectiveness on different backbones, we implement various backbones as shown in Tab. \ref{tab:diffbackbone}. The experiment result indicates that our method is equally valid for different backbones. Even for a backbone strong enough, our method can still improve the performance. We report the best result of ConvNext-L \cite{liu2022convnet} as backbone which achieves 34.57 weighted mIoU. In the OpenEarthMap Land Cover Mapping Few-Shot Challenge, the result ranked No.2 till we finished the paper.

\subsection{Visualization of transition matrix} \label{sec:vis_trans}

\begin{figure}[t]
        \centering
        \includegraphics[width=6.0cm]{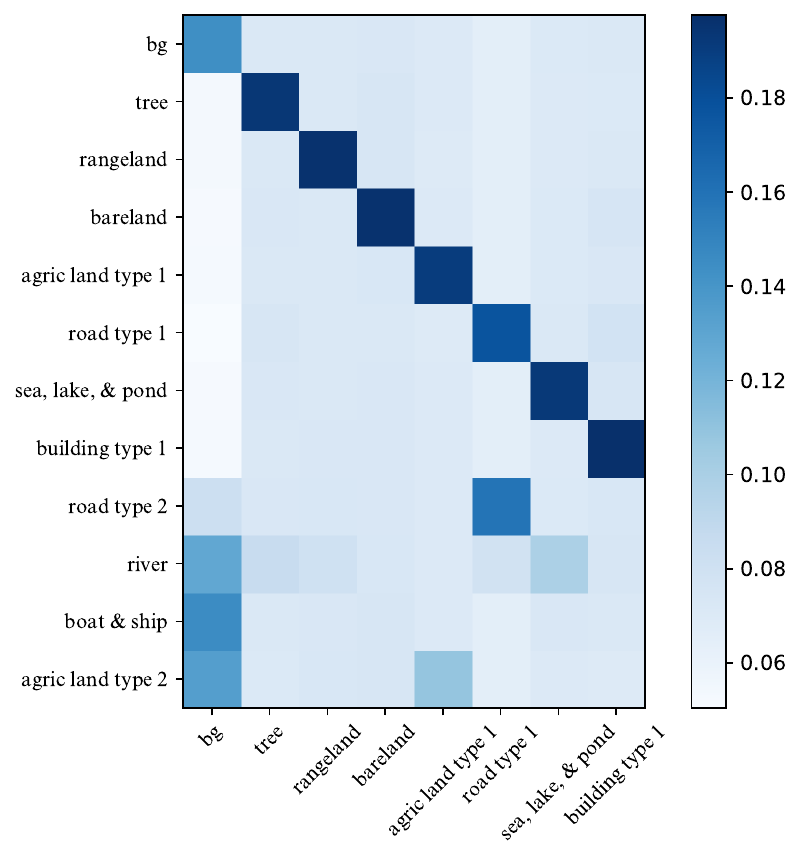}
        \caption{\textbf{The heat map of the Transition Matrix after the few-shot learning phase}.}
        \label{figure:heatmap}
\end{figure}

In our method, we do not leverage the knowledge distillation to perverse the knowledge learned from base classes. Instead, we achieve the preservation by extending the transition matrix with base-to-base parts. To verify our claim and show the effectiveness of base-to-base parts, we visualize the transition matrix, as shown in Fig. \ref{figure:heatmap}.

% 验证有效性还需要加一个实验，就是transition matrix不要base to base 的部分，直接就是8-4
As Fig. \ref{figure:heatmap} indicates, the diagonal elements of the base-to-base parts, $ \mathbf{S}^{\text{base2base}}(\mathbf{x})$, are dominant in the corresponding columns which suggests that after the outer product with classification probabilities given by $W^b_t$, the logit on the corresponding position is much greater than others. After the softmax, the base classes have greater logits. Thus, our proposed method succeeds in preserving the knowledge from base classes.

Besides, the first column in the transition matrix represents the similarity transition probabilities of the background class in the training set to base and novel classes during the few-shot learning.
It can be seen that the transition probabilities from background to novel classes stand high while background to base classes are low. This is because, during the training phase, potential novel classes are treated as background whereas base classes are not. Thus, having high transition probabilities from background to novel classes, the transition branches further help the model distinguish the classes that were previously considered as background in the training set.

\subsection{Quantitative study of overfitting}

As we claim optimizing the classifier on the support set may result in overfitting. We do a quantitative study to verify our claim, as shown in Fig. \ref{fig:overfit}. \textbf{A)} Obviously, as shown in Fig. \ref{fig:overfit1}, as the optimization keeps, the classifier suffers from overfitting, i.e., the mIoU on the support set kept increasing while that of query images started decreasing as it reached the highest score at about 450 epoch. \textbf{B)} Compared to $k=1$ and $k=4$, $\mathcal{L}_\pi$ significantly delays the onset of overfitting from about 450 epochs to about 620 epochs. Moreover, the classifier achieves higher performance on query images. The performance without $\mathcal{L}_\pi$ is below 4, whereas with it, the performance exceeds 5. \textbf{C)} Compared $k=4$ to $k=1$, the classifier keeps gaining improvement on the query image and does not reach the pike yet, suggesting our method still has ample untapped potential, enabling higher performance to be achieved.

\subsection{Ablation study}

To validate our proposed method, we do various ablation studies from different perspectives.

\paragraph{Ablation on components.} To verify the effectiveness of our components we do ablation studies on all of our components. The results show that: \textbf{a)} Without $\mathcal{L}_\pi$, the model achieves better results on base classes but shows a degradation on novel classes, corresponding to the previous experiment that reveals the overfitting happened on the support set. \textbf{b)} Without transition logits, the model shows a degradation in novel classes, especially on Argic land type 2, proving our Similarity transition is valid yet effective. \textbf{c)} Without $\mathcal{L}_{\textit{LDAM}}$, the model shows great degradation in novel classes for the disproportionate number of samples of base classes to novel classes. \textbf{D)} With all the components, our method achieves a significant performance improvement on novel classes with an accepting degradation on old classes. In summary, with all components armored, our method improves the performances prominently in novel classes with a minor decrease in base classes.

\begin{table}[htp]
\centering
\caption{\textbf{The ablation study of our method.} We remove  $\text{Logits}_{tr}$ and $\mathcal{L}_\textit{LDAM}$, \ie similarity utilization and class balancing.}
\label{tab:ablation}

\resizebox{0.40\textwidth}{!}{%

\begin{tabular}{l c c c }
\hline
     & w/o $\text{Logits}_{tr}$ &  w/o $\mathcal{L}_{\textit{LDAM}}$& ours \\ 
\hline
 Tree&\textbf{60.26}& \textbf{60.26}& 60.22\\
Rangeland&59.33& \textbf{59.50}& 59.32\\
Bareland&\textbf{35.99}& 35.04& 35.98\\
Agric land type 1&75.47& \textbf{75.51}& 75.47\\
 Road type 1&\textbf{55.12}& 54.55&55.06\\
 Sea, lake, \& pond&40.26& 40.28& \textbf{40.29}\\
 Building type 1&61.80& 61.78& \textbf{61.86}\\
 \hline
 Road type 2&\textbf{0.44}& 0.35&\textbf{0.44}\\
 River&\textbf{39.98}& 30.14&39.13\\
 Boat \& ship&0.00& 0.00&0.00 \\
 Agric land type 2&41.16& 29.70&\textbf{47.28}\\
   \hline
 Base mIoU&\textbf{55.46}& 55.28&\textbf{55.46}\\
 Novel mIoU&20.17& 15.05&\textbf{21.71}\\
 Average mIoU&37.82& 35.16& \textbf{38.58}\\
  \hline
 weighted mIoU&34.29& 31.14&\textbf{35.21}\\
   \hline
\end{tabular}%
}
\end{table}

\paragraph{Ablation on the backbone compared with DIaM \cite{diam}.} To prevent our proposed method overfits a specific backbone, we do an ablation study on different backbones, proving the generalization of our methods. In detail, we compared our method with DIaM \cite{diam} with different backbones, i.e. ResNet-101 with PSPNet \cite{zhao2017pyramid}, ViT-B/16 \cite{dosovitskiy2020image} with UperNet \cite{xiao2018unified}, and ConvNext-L with UperNet. \textbf{A)} Compared with DIaM, except for ResNet-101 with PSPNet, our method beats DIaM in all classes with all backbones. This reveals that our method works better on GFSS than DIaM. \textbf{B)} Compared to our method with different backbones, the performance gaining is steady, \ie 3.62, 2.39, 3.60 respectively. This suggests that our method, as a plug-in module to enable any segmentation model owning the few-shot learning ability, can continually gain improvements, indicating with a stronger backbone implemented, our method can achieve even better performance. Unluckily, due to the limited time, we did not implement more backbones to prove our methods' effectiveness in a more general field. 

% To have a straightforward understanding of the problems we stressed, qualitative research is presented. \textbf{A)} Without distillation loss, Catastrophic forgetting phenomena occur. It is evidenced in the third column of the figures, where base classes such as 'Building type 1', 'Agric land type 1', and 'Sea, Lake, \& pond' are mislabeled, while there is an evident improvement in the fourth column. Introducing LDAM loss balances the issue of uneven distribution, resulting in increased confidence in the new classes, as demonstrated by the expansion of 'Agric land type 2' and 'River' into reasonable regions.

\begin{table}[htp]
\centering
\caption{\textbf{Weighted mIoU in different backbones compared with DIaM \cite{diam}.} Performance on the left side of the slash refers to the results of DIaM \cite{diam}, ours on the right side.}
\label{tab:diffbackbone}
\resizebox{0.45\textwidth}{!}{%

\begin{tabular}{lccccc}
\hline
     Backbone& ResNet-101&ViT-B/16& ConvNext-L\\
 Seg.Head &PSPNet&UPerNet &UPerNet\\
   \hline
 Base mIoU& \textbf{37.41} / 37.37 & 46.48 / \textbf{48.08} & 54.78 / \textbf{55.46}\\
 Novel mIoU& 4.13 / \textbf{10.19} & 12.33 / \textbf{15.24} & 16.16 / \textbf{21.71}\\
 Average mIoU& 20.77 / \textbf{23.78} & 29.41 / \textbf{31.66} & 35.47 / \textbf{38.58}\\
  \hline
 weighted-mIoU& 17.44 / \textbf{21.06} & 25.99 / \textbf{28.38} & 31.61 / \textbf{35.21}\\
   \hline
\end{tabular}%
}
\end{table}

% \begin{table*}[!htbp]
% \centering
% \caption{Results}
% \tiny
% \label{table6}
% \resizebox{1\textwidth}{!}{%
% \centering
% \begin{tabular}{l c c c c c c cllll}
% \hline
%             & Tree& Rangeland& Bareland& Agric land type 1& Road type 1& Sea, lake, \& pond& Building type 1 & Road type 2& River& Boat \& ship&Agric land type 2\\ 
% \hline
% no distillation& & & &  & & &  & & & &\\
% no similarity& &  &  & & & &   & & & &\\
% no class imbalance& & & & & & &  & & & &\\  
% \hline
%  ours& & && && &  & & & &\\  \hline
% \end{tabular}%
% }
% \end{table*}

\section{Conclusions and Discussions}

In this work, we determine two issues that exist in the GFSS task and verify our claim with experiments. Our key insight is to utilize the similarity between the base classes and novel classes as well as processing the class imbalanced in the training set and support set. Beyond that, we conduct plenty of experiments to show the effectiveness of our methods, demonstrating that our proposed components address the issues we found. In the future, we will further investigate how to encode the similarity more effectively. Since we do not obtain any supervision from the class similarities, we train the model in an end-to-end manner, hoping the model learns the similarities implicitly. However, with extra supervision provided well-designed architectures are meant to improve the performance. Furthermore, other valid methods to prevent overfitting the support set need to be studied.
{
    \small
    \bibliographystyle{ieeenat_fullname}
    \bibliography{main}
}

% WARNING: do not forget to delete the supplementary pages from your submission 
% \input{sec/X_suppl}

\end{document}